\documentclass[10pt,twocolumn,letterpaper]{article}

\usepackage{cvpr}              
\usepackage[export]{adjustbox}
\usepackage{booktabs}        
\usepackage{amsmath}         
\usepackage{multirow}        
\usepackage{array}           
\usepackage{caption}         
\usepackage{siunitx}         
\usepackage{amsmath}

\newcommand{\best}[1]{\textbf{#1}}
\definecolor{cvprblue}{rgb}{0.21,0.49,0.74}
\usepackage[pagebackref,breaklinks,colorlinks,allcolors=cvprblue]{hyperref}
\usepackage{caption}
\usepackage{booktabs}
\usepackage{siunitx}
\def\confName{CVPR}
\def\confYear{2026}

\title{HiDiGen: Hierarchical Diffusion for B-Rep Generation \\ with Explicit Topological Constraints}

\author{
    Shurui Liu$^{1}$ \quad 
    Weide Chen$^{2}$ \quad 
    Ancong Wu$^{1,}$\thanks{Corresponding author} \\
    $^{1}$School of Computer Science and Engineering, Sun Yat-sen University, China \\
    $^{2}$School of Intelligent Systems Engineering, Shenzhen Campus of Sun Yat-sen University, China \\
    {\tt\small \{liushr29, chenwd56\}@mail2.sysu.edu.cn, wuanc@mail.sysu.edu.cn}
}

\begin{document}
\maketitle
\begin{abstract}
Boundary representation (B-rep) is the standard 3D modeling format in CAD systems, encoding both geometric primitives and topological connectivity. Despite its prevalence, deep generative modeling of valid B-rep structures remains challenging due to the intricate interplay between discrete topology and continuous geometry. In this paper, we propose \textbf{HiDiGen}, a hierarchical generation framework that decouples geometry modeling into two stages, each guided by explicitly modeled topological constraints. Specifically, our approach first establishes face-edge incidence relations to define a coherent topological scaffold,  upon which face proxies and initial edge curves are generated.  Subsequently, multiple Transformer-based diffusion modules are employed to refine the geometry by generating precise face surfaces and vertex positions, with edge-vertex adjacencies dynamically established and enforced to preserve structural consistency. This progressive geometry hierarchy enables the generation of more novel and diverse shapes, while two-stage topological modeling ensures high validity.  Experimental results show that \textbf{HiDiGen} achieves strong performance, generating novel, diverse, and topologically sound CAD models.

\end{abstract}    
\section{Introduction}
\begin{figure}
    \centering
    \includegraphics[width=\linewidth, margin=0pt 0pt 100pt 30pt]{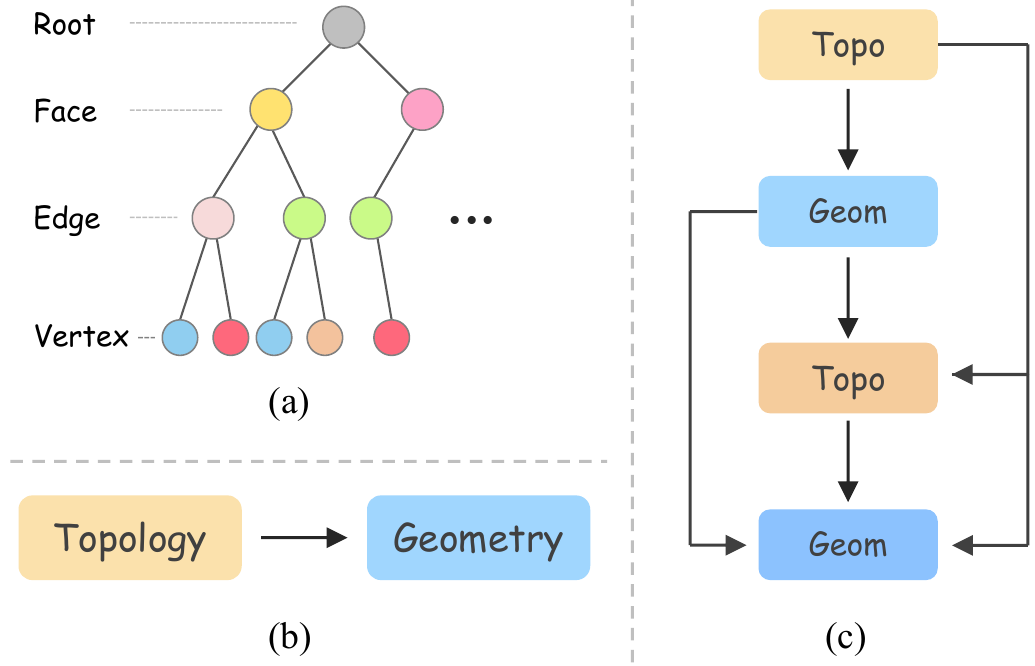}
    \caption{\textbf{Comparison of learning paradigms for B-rep generation}: (a) implicit topology encoding, (b) decoupled topology-geometry generation, and (c) our hierarchical generation framework.}
    \label{fig:paradiam}
\end{figure}
\label{sec:intro}
Boundary representation (B-rep)~\cite{weiler1986topological} is the de facto standard for 3D shape modeling in Computer-Aided Design (CAD), offering a precise, structured description of solid objects through parametric surfaces (faces), curves (edges), and vertices. 
Unlike mesh-based approximations composed of planar facets, B-reps explicitly encode both high-fidelity geometry and rich topological connectivity, enabling accurate modeling of complex free-form solids—making them indispensable in engineering design and manufacturing pipelines.

However, generating valid B-rep models automatically remains a challenging task due to the intricate coupling between geometric accuracy and topological consistency. 
Existing generative approaches often bypass this complexity by predicting sequences of constructive solid geometry (CSG) or sketch-based operations (e.g., extrusion, revolution)~\cite{wu2021deepcad, xu2022skexgen, xu2023hierarchical}, relying on off-the-shelf CAD kernels to reconstruct B-reps in post-processing. 
While effective for simple primitives, these methods are typically constrained by limited dataset scales~\cite{willis2021fusion, wu2021deepcad} and restricted to a narrow set of procedural rules, limiting their applicability to real-world, complex designs.

Recent advances have explored direct B-rep generation, with notable progress demonstrated by methods such as BrepGen~\cite{xu2024brepgen} and DTGbrepGen~\cite{li2025dtgbrepgen}. However, these approaches lack explicit mechanisms to model the mutual dependencies between geometry and topology. As illustrated in Fig.~\ref{fig:paradiam}, BrepGen implicitly embeds topological information within the geometry generation process, which often leads to incomplete or distorted surface reconstructions.

In contrast, DTGbrepGen decouples topology and geometry prediction but omits geometric feedback during topological reasoning, leading to invalid or inconsistent edge-face connectivities. 
These limitations underscore a critical gap: the lack of a unified, hierarchical framework for jointly modeling geometry and topology, which often results in inconsistent topological structures and geometric representations, leading to artifacts such as non-manifold edges, degenerate faces, and other structural irregularities.

\begin{figure*}[htbp]
    \makebox[\textwidth][c]{
        \includegraphics[width=1.0\textwidth]{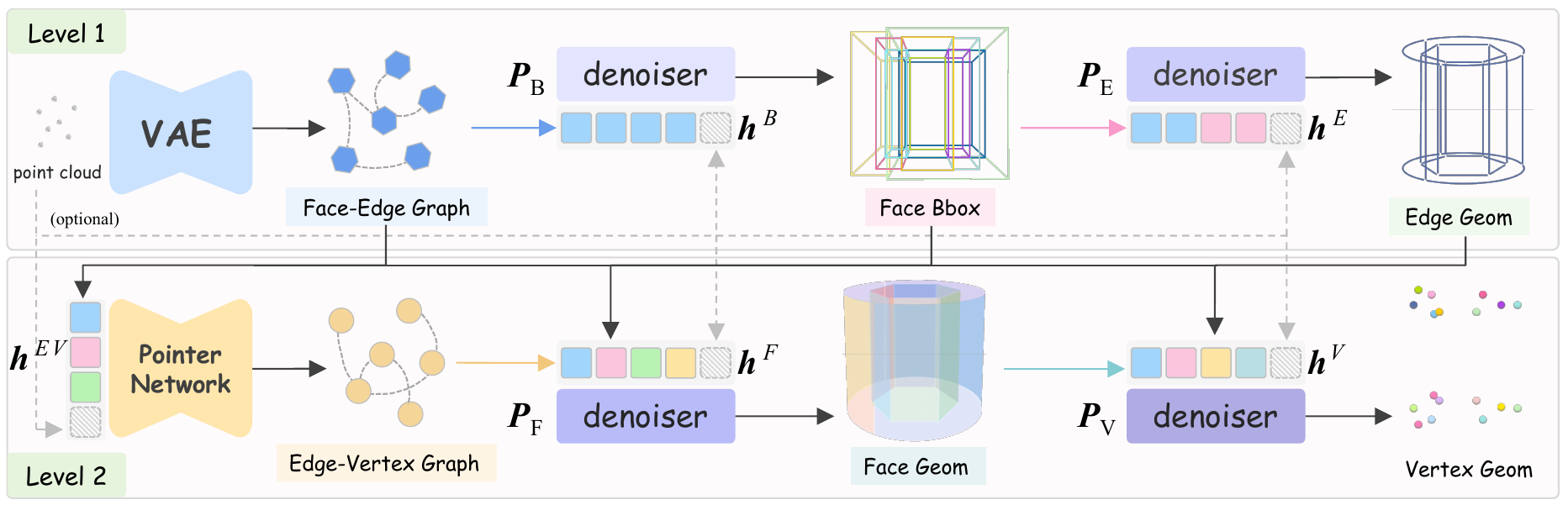}%
    }
\caption{\textbf{HiDiGen pipeline overview.} Our framework generates B-rep models in two hierarchical levels: Level~1 produces face-edge topology ($\mathbf{EF}_i$) and coarse geometry (face bounding boxes $\mathbf{B}_i$, edge curves $\mathbf{E}_i$); Level~2 refines edge-vertex topology ($\mathbf{EV}_i$) and detailed geometry (face surfaces $\mathbf{F}_i$, vertex positions $\mathbf{V}_i$), with each stage conditioned on prior outputs.}
    \label{fig:pipeline}
\end{figure*}

To address this challenge, we propose \textbf{HiDiGen}, a \textbf{Hierarchical} and \textbf{Di}rectional \textbf{Gen}erative model for B-rep construction. 
HiDiGen introduces a two-stage, topology-guided generation paradigm that enables bidirectional interaction between geometric details and topological structure. 
In the first stage, we generate a valid edge-face adjacency graph, which serves as a structural prior for subsequent face proxy and coarse edge geometry synthesis. 
Conditioned on this intermediate representation, the second stage refines the topology via edge-vertex adjacencies and drives the generation of final face geometries and precise vertex positions. 
By decomposing the generation process into progressively finer topological constraints, HiDiGen ensures both geometric realism and topological validity at each level. Our key contributions are threefold:

\begin{itemize}
\item We propose HiDiGen, a hierarchical  generative framework for B-rep modeling that explicitly decouples geometry and topology generation into two coherent stages. Our approach introduces a topology-guided paradigm with bidirectional interaction, enabling structured priors to inform geometric synthesis while allowing geometric feedback to refine topological consistency.

    \item We design an efficient intermediate fusion strategy that enables effective propagation and cross-stage integration of topological and geometric cues.
    \item Extensive experiments demonstrate that HiDiGen generates 3D B-rep models with higher topological validity, geometric accuracy, and shape diversity compared to existing methods.
\end{itemize}

\section{Related Work}
Generating CAD models has attracted growing attention, driven by advancements in geometric deep learning and 3D representation learning. A diverse set of methods has been proposed to handle various CAD representations, ranging from parametric sketches to boundary representations. This section provides an overview of the main methodologies.
\subsection{Constructive Solid Geometry (CSG)}
Constructive Solid Geometry (CSG) is a widely used representation for 3D shapes, where primitives(e.g., cuboids, spheres) are combined via Boolean operations (e.g., union, subtraction) to form a hierarchical CSG tree. Due to its compositional structure and interpretability, CSG has been extensively adopted in shape program synthesis~\cite{NEURIPS2019_50d2d226, tian2019learning}. While CSG representations can in principle be converted to B-reps, such conversions often introduce topological degeneracies, particularly in near-coincident configurations, leading to fragile and non-manifold structures. In contrast, our approach directly learns from high-quality B-rep topologies prevalent in industrial CAD models, enabling the generation of robust outputs compatible with standard direct modeling tools.

\subsection{Command Sequence}
Sketch and extrude is a widely adopted paradigm in parametric CAD modeling, where a 3D model is constructed as a sequence of operations typically involving sketching 2D profiles followed by extrusion, encoded in a parametric CAD file. DeepCAD~\cite{wu2021deepcad} introduced a large-scale dataset of such sequential CAD modeling commands, establishing a foundational benchmark that has spurred significant advances in sequence-based CAD understanding and generation~\cite{xu2022skexgen,xu2023hierarchical,chen2025cadcrafter}. While recent methods have considerably improved the fidelity and controllability of generated designs, existing sequential generative models remain restricted to sketches composed solely of line, arc, and circle primitives, coupled exclusively with the extrude operation. In contrast, our approach directly synthesizes boundary representation (B-rep) geometry, with a focus on enabling complex curves and freeform surfaces that are beyond the expressive capacity of prior sketch and extrude frameworks.

 \subsection{Boundary Representation}
B-rep models define solid geometry through vertices, edges, and faces, offering a precise and structured representation that is fundamental to capturing complex surface interactions in industrial design and manufacturing workflows. Prior works have largely focused on  tasks such as classification and segmentation~\cite{cao2020graph,willis2022joinable}. Recent approaches extend to the generative modeling of B-rep structures. For instance, SolidGen~\cite{jayaraman2022solidgen} constructs B-rep models via autoregressive generation, beginning with vertices and conditionally predicting edges and faces; however, it is restricted to a limited class of shapes. BrepGen~\cite{xu2024brepgen} formulates B-rep generation as a tree-structured process, leveraging multiple diffusion models to generate geometric attributes while implicitly encoding topological relationships through the hierarchical tree structure. In contrast, DTGbrepGen~\cite{li2025dtgbrepgen} explicitly decouples topology and geometry into two separate generation stages. Similarly, Brepdiff~\cite{brepdiff} relies on geometry-driven post-processing to recover topological consistency. Hola~\cite{liu2025hola} unifies geometric and topological learning by encoding both surfaces and their intersecting curves into a single holistic surface latent space, where topological connections are inferred through a neural intersection module that reconstructs curve geometry from surface pairs.  Despite their strong empirical performance, none of these methods explicitly model the interaction between topological and geometric information during the generative process, nor do they employ bidirectional supervision to enforce mutual consistency between the two domains.
\subsection{Generate CAD Models from Point Cloud}
Reconstructing CAD models from point clouds is a fundamental problem in reverse engineering. Traditional pipelines first convert point clouds into meshes before extracting CAD representations~\cite{benkHo2004segmentation}, whereas recent learning-based approaches directly predict parametric CAD structures without relying on intermediate geometry. Notable examples include Point2Cyl, which decomposes shapes into extruded cylindrical primitives~\cite{uy2022point2cyl}, and ComplexGen, which generates full B-rep models by inferring geometric primitives and their topological relationships ~\cite{guo2022complexgen}. Despite this progress, accurate CAD reconstruction remains challenging due to incomplete observations, high geometric complexity, and the inherent ambiguity in recovering precise topological configurations.

\section{Hierarchical Generation Framework}
\label{sec: geometry_generation}

In this section, we will outline the basic idea of our hierarchical generation framework.
\subsection{Problem Setup}
\label{sec: problem_setup}

Let $\mathbf{M}_i$ represent a valid 3D solid modeled using a boundary representation (B-rep), defined as $\mathbf{M}_i = (\mathcal{T}_i, \mathcal{G}_i)$. Here, $\mathcal{T}_i$ encodes the combinatorial topology through incidence and adjacency relationships among vertices, edges, and faces, while $\mathcal{G}_i$ provides the geometric attributes.

We employ B-spline parameterizations to represent all geometric primitives. 
Let $N_i^v$, $N_i^e$, and $N_i^f$ denote the number of vertices, edges, and faces, respectively, in the $i$-th shape.
Vertices are stored as 3D Cartesian coordinates in the vertex tensor $\mathbf{V}_i \in \mathbb{R}^{N_i^v \times 3}$.
Each edge is modeled as a cubic B-spline curve defined by four 3D control points; accordingly, the edge geometry is encoded in $\mathbf{E}_i \in \mathbb{R}^{N_i^e \times 12}$, where each row concatenates the coordinates of the four control points.
Faces are represented as bi-cubic B-spline surfaces over a $4 \times 4$ control grid, yielding the face geometry tensor $\mathbf{F}_i \in \mathbb{R}^{N_i^f \times 48}$.
For each face, we also compute its axis-aligned bounding box, stored in $\mathbf{B}_i \in \mathbb{R}^{N_i^f \times 6}$.
The complete geometric realization of the $i$-th shape is then given by $\mathcal{G}_i = \{\mathbf{B}_i, \mathbf{F}_i, \mathbf{E}_i, \mathbf{V}_i\}.$

The topological structure $\mathcal{T}_i$ is represented through incidence relations. The face-edge incidence matrix $\mathbf{EF}_i \in \mathbb{N}^{N_e^i \times 2}$ records, for each edge, the indices of its two adjacent faces. The edge-vertex incidence matrix $\mathbf{EV}_i \in \mathbb{N}^{N_e^i \times 2}$ stores the indices of the two vertices connected by each edge. Thus, the topological realization is $\mathcal{T}_i = \{\mathbf{EF}_i, \mathbf{EV}_i\}$.

To facilitate learning,  we decouple the generative process into hierarchical stages. We first redefine the geometry into coarse and fine components: $\mathcal{G}_{i,1} = \{\mathbf{B}_i, \mathbf{E}_i\}$ denotes the coarse geometric information, while $\mathcal{G}_{i,2} = \{\mathbf{F}_i, \mathbf{V}_i\}$ contains the detailed geometric data. Similarly, the topology is split into $\mathcal{T}_{i,1} = \{\mathbf{EF}_i\}$, governing global face-edge connectivity, and $\mathcal{T}_{i,2} = \{\mathbf{EV}_i\}$, encoding local edge-vertex connections.

Our goal is to learn a generative model that synthesizes B-rep structures $\hat{\mathbf{M}} = (\hat{\mathcal{T}}, \hat{\mathcal{G}})$ that are both topologically consistent and geometrically plausible. To this end, we factorize the joint distribution as follows:

\begin{equation}
\begin{aligned}
p(\hat{\mathcal{T}}_1, \hat{\mathcal{G}}_1, \hat{\mathcal{T}}_2, \hat{\mathcal{G}}_2) = 
&\underbrace{p(\hat{\mathcal{T}}_1)}_{\text{Level 1: Topology}} 
\cdot \underbrace{p(\hat{\mathcal{G}}_1 \mid \hat{\mathcal{T}}_1)}_{\text{Level 1: Geometry}} \\
\cdot &\underbrace{p(\hat{\mathcal{T}}_2 \mid \hat{\mathcal{T}}_1, \hat{\mathcal{G}}_1)}_{\text{Level 2: Topology}} 
\cdot \underbrace{p(\hat{\mathcal{G}}_2 \mid \hat{\mathcal{T}}_1, \hat{\mathcal{G}}_1, \hat{\mathcal{T}}_2)}_{\text{Level 2: Geometry}}
\end{aligned}
\end{equation}

\subsection{Method Overview}
A key challenge in B-reps models lies in the intricate interplay between topology and geometry. To address this, HiDiGen decouples the geometric modeling process into two distinct stages, each governed by learned topological constraints. This hierarchical formulation enables structured control, where topology guides geometry, while refined geometric outputs in turn inform and improve the topological structure, fostering mutual enhancement. As illustrated in Figure~\ref{fig:pipeline}, at \textit{Level 1}, we generate the global face-edge adjacency structure $\mathbf{EF}_i$, establishing the high-level topological skeleton of the shape. This topology then conditions the subsequent generation of coarse geometric proxies $\mathbf{B}_i$, followed by edge geometry $\mathbf{E}_i$, ensuring initial alignment between adjacent faces.

At \textit{Level 2}, conditioned on the previously generated $\mathbf{EF}_i$ and intermediate geometries, we refine the topology further by generating the edge-vertex incidence $\mathbf{EV}_i$, which defines how edges connect at vertices. This finer-grained connectivity structure then guides the generation of detailed face geometry $\mathbf{F}_i$ and vertex positions $\mathbf{V}_i$, enforcing local geometric consistency across shared boundaries.

\subsection{Contextual Conditioned Generation}

As illustrated in Fig.~\ref{fig:pipeline}, during the hierarchical generation process, each stage encodes attributes produced by the preceding stage, enabling progressively richer conditional representations. Consequently, higher level generation steps benefit from increasingly contextualized information. In this section, we present a detailed analysis of the feature representation dynamics throughout the generation pipeline.

As mentioned in Sec.\ref{sec: problem_setup}, the task can be reformulated as:
\begin{equation}
\begin{aligned}
p(\hat{\mathcal{T}}_1, \hat{\mathcal{G}}_1, \hat{\mathcal{T}}_2, \hat{\mathcal{G}}_2) 
= &\, p(\hat{\mathcal{T}}_1 \mid \mathbf{h}^\emptyset) \\
  &\cdot p(\hat{\mathcal{G}}_1 \mid \hat{\mathcal{T}}_1; \, \mathbf{h}^{(1)}) \\
  &\cdot p(\hat{\mathcal{T}}_2 \mid \hat{\mathcal{T}}_1, \hat{\mathcal{G}}_1; \, \mathbf{h}^{(2)}) \\
  &\cdot p(\hat{\mathcal{G}}_2 \mid \hat{\mathcal{T}}_1, \hat{\mathcal{G}}_1, \hat{\mathcal{T}}_2; \, \mathbf{h}^{(3)}),
\end{aligned}
\end{equation}
\begin{figure}[htbp]
    \makebox[\linewidth]{
        \includegraphics[width=1.0\linewidth]{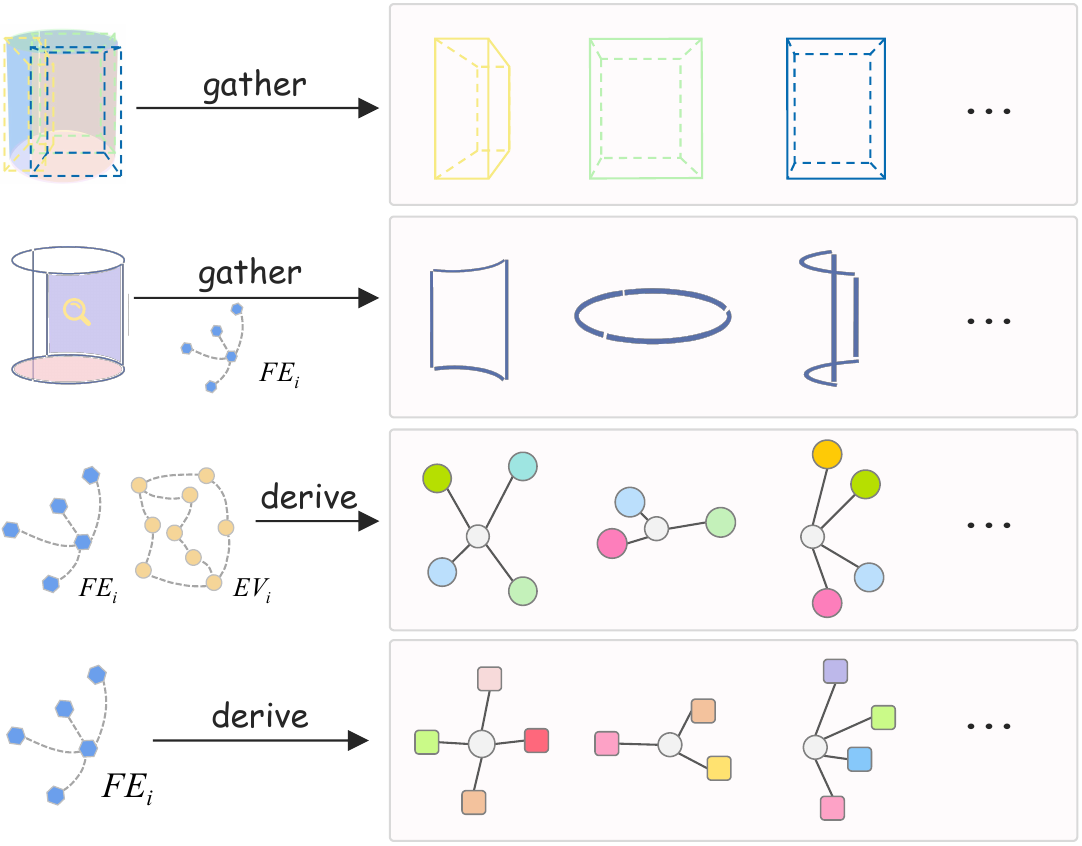}%
    }
    \caption{\textbf{Illustration of the contextual conditioned embedding for face geometry generation.} We gather the corresponding face bounding boxes and edge geometries, and further derive topological connectivity relationships to form the conditioning embedding.}
    \label{fig:story}
\end{figure}
where each contextual embedding $\mathbf{h}^{(k)}$ is deterministically computed from the previously generated components: 
$\mathbf{h}^{(1)} = \Phi_1(\hat{\mathcal{T}}_1)$ encodes face and edge connection topology context; 
$\mathbf{h}^{(2)} = \Phi_2(\hat{\mathcal{T}}_1, \hat{\mathcal{G}}_1)$ captures Level 1 geometry and topology context; 
and $\mathbf{h}^{(3)} = \Phi_3(\hat{\mathcal{T}}_1, \hat{\mathcal{G}}_1, \hat{\mathcal{T}}_2)$ aggregates fine grained face and vertex conditions. In the following, we introduce the complete modeling process.

\subsubsection*{Level-1 Topology Generation}

At this stage, our goal is to learn the distribution over Level-1 topology, denoted as $p(\hat{\mathcal{T}}_1)$. Formally, we define $\mathcal{T}_{1,i} = \mathbf{EF}_i$, where $\mathbf{EF}_i$ represents the extracted topological features for the $i$-th instance.As diclaimed in DTGbrepgen~\cite{li2025dtgbrepgen}, directly learning the distribution of edge-face matrices ${\mathbf{EF}_i}$ is suboptimal. We use a invariant representation $\mathbf{FeF}_i$ as same as DTGBrepgen~\cite{li2025dtgbrepgen}, where each $\mathbf{FeF}_i \in \mathbb{N}^{N_i^f \times N_i^f}$ is a symmetric matrix such that the entry $\mathbf{FeF}_i[k,l]$ denotes the number of shared edges between face $k$ and face $l$ in the B-rep model $\mathbf{M}_i$.  And we standardize the face indexing by sorting faces in ascending order according to their number of boundary edges. We extract the upper triangular entries of $\mathbf{FeF}_i $ (excluding the diagonal) in row-major order and flatten them into a sequence:
$ \mathbf{EF}_i^{\text{seq}} := \left( \mathbf{FeF}_i[p,q] \right)_{1 \leq p < q \leq N_f^i} $,
where $ p $ and $ q $ denote face indices satisfying $ p < q $, ensuring each pair is included exactly once. $\mathbf{FeF}_i$ serves as a canonical, topology-preserving signature of the B-rep structure and is used as input to our generative model.

In the $\mathbf{EF}_i$ stage, we first encode the edge feature sequence $\mathbf{EF}_i^{\mathrm{seq}}$ into a latent representation using a Transformer encoder . A masked Transformer decoder~\cite{vaswani2017attention} then autoregressively reconstructs $\mathbf{EF}_i^{\mathrm{seq}}$ from this latent space. To encourage structured and disentangled representations, we formulate this process as a variational autoencoder (VAE)~\cite{kingma2013auto}, optimizing the standard VAE objective that combines reconstruction loss and KL regularization between the learned posterior and a prior distribution in the latent space.

\subsubsection*{Level-1 Geometry Generation}
The goal of this stage is to model the Level-1 geometry distribution $p(\hat{\mathcal{G}}_1 \mid \hat{\mathcal{T}}_1)$, where each geometric primitive $\mathcal{G}_{1,i} = \{\mathbf{B}_i, \mathbf{E}_i\}$ consists of a face bounding box $\mathbf{B}_i$ and associated edge geometries $\mathbf{E}_i$. To this end, we introduce a hierarchical generative framework for 3D shape synthesis that produces geometric components in a structured and sequential manner, conditioned on topological and geometric priors inferred in preceding stages. Our method employs Transformer-based diffusion models~\cite{dhariwal2021diffusion,ho2020denoising,vaswani2017attention} to parameterize the target distribution. Specifically, we design two dedicated denoising networks: $\mathcal{P}_B$ for generating face bounding boxes $\mathbf{B}_i$, and $\mathcal{P}_E$ for edge geometries $\mathbf{E}_i$.

The $\mathbf{B}_i$ generation pipeline is initialized by constructing per-face embeddings that integrate both combinatorial features from prior inference stages. For face bounding box prediction, the initial embedding is defined as $ \mathbf{h}_i^B = \mathbf{FE}_i $, where $ \mathbf{FE}_i $ denotes the transpose of the edge-to-face incidence matrix $ \mathbf{EF}_i $. The matrix $ \mathbf{EF}_i $, obtained from the preceding topology inference stage, encodes the set of edges bounding each face and thus captures the underlying combinatorial structure of the shape. The denoising network $\mathcal{P}_B$ takes this embedding together with noisy bounding box inputs and iteratively refines them through the diffusion process:

\begin{equation}
    \mathbf{B}_i = \mathcal{P}_B\big(\boldsymbol{\epsilon}_T;\, \mathbf{h}_i^B\big), \quad \text{where } \boldsymbol{\epsilon}_T \sim \mathcal{N}(\mathbf{0}, \mathbf{I}).
\end{equation}

Given the predicted face bounding boxes $ \mathbf{B}_i $ and the incidence $ \mathbf{EF}_i $, we proceed to generate the corresponding edge geometries $ \mathbf{E}_i $. To this end, we design a contextual gather module that aggregates bounding box information from incident faces while explicitly preserving edge-face connectivity. Specifically, we apply a learnable function $ \Phi_{E} $ that first encodes the gathered features and the incidence structure into a joint representation, then performs summation over incident elements to yield a compact edge-centric embedding:
\begin{equation}
\mathbf{h}_i^{E} = \Phi_{E} \left( \operatorname{\mathsf{gather}}(\mathbf{EF}_i, \mathbf{B}_i), \mathbf{EF}_i \right),
\end{equation}
where $ {\operatorname{\mathsf{gather}}}(\mathbf{B}_i, \mathbf{EF}_i) $ retrieves the bounding boxes of faces incident to each edge based on the connectivity specified in $ \mathbf{EF}_i $. This edge-centric embedding serves as the conditioning input to the edge denoising network $\mathcal{P}_E$, which generates the final edge geometries $\mathbf{E}_i$. By explicitly incorporating $ \mathbf{EF}_i $ into the input of $ \mathbf{h}_E $, the model can directly reason about incidence relationships between edges and faces, thereby enhancing structural coherence in the generated geometries. Ablation studies verify that explicitly injecting topological structure significantly improves generation performance, confirming its critical role in guiding geometric synthesis. We employs a same standarized diffusion process for $\mathbf{h}_E$:

\begin{equation}
    \mathbf{E}_i = \mathcal{P}_E\big(\boldsymbol{\epsilon}_T;\, \mathbf{h}_i^E\big), \quad \text{where } \boldsymbol{\epsilon}_T \sim \mathcal{N}(\mathbf{0}, \mathbf{I}).
\end{equation}

\subsubsection*{Level-2 Topology Generation}
We aim to model the Level-2 topology distribution $p(\hat{\mathcal{T}}_2 \mid \hat{\mathcal{T}}_1, \hat{\mathcal{G}}_1)$, following the directed half-edge formulation introduced in DTGBrepGen~\cite{li2025dtgbrepgen} to ensure unambiguous vertex-wise connectivity. Each undirected edge is decomposed into two oppositely directed half-edges, enabling consistent traversal along face boundaries. To achieve deterministic serialization, we assign each edge a unique global identifier by lexicographically sorting its incident face indices $(f_a, f_b)$ with $f_a < f_b$, resolving ties arbitrarily in non-manifold cases. For each face, the boundary sequence begins with the half-edge corresponding to the smallest global ID and proceeds in a fixed winding order (e.g., counter-clockwise), resulting in a canonical and reversible encoding of the boundary representation.

In this process, edge-vertex adjacency generation is reformulated as a sequence modeling task. Specifically, we serialize the half-edge connections within each face in canonical order, yielding an ordered edge--vertex sequence for body $\mathbf{M}_i$:
\begin{equation}
    \mathbf{EV}_i^{\mathrm{seq}} := [ s_{i}^{j,1}, s_{i}^{j,2}, \ldots, \langle \mathrm{L} \rangle, \langle \mathrm{F} \rangle, \ldots, \langle \mathrm{F} \rangle ],
    \label{eq:evseq}
\end{equation}
where $s_{i}^{j,k}$ denotes the $k$-th half-edge in the $j$-th face, and $\langle \mathrm{L} \rangle$, $\langle \mathrm{F} \rangle$ are special tokens indicating the end of a loop and the end of a face, respectively. This sequential structure explicitly encodes topological continuity: consecutive half-edges in the sequence share a common vertex, facilitating autoregressive modeling of facial boundary structures.

Our approach integrates geometric and topological priors into the sequence generation process. During edge--vertex adjacency prediction, we utilize available structural and geometric information, including the edge-face incidence $\mathbf{EF}_i$, face bounding boxes $\mathbf{B}_i$, and per-edge geometric embeddings $\mathbf{E}_i$. We employ a fusion function $\Phi_{ev}$ together with a $\operatorname{gather}$ operation to encode these multimodal inputs into a unified embedding as same as others generation process:
\begin{equation}
    \mathbf{h}_i^{EV} = \Phi_{EV}\left( \operatorname{\mathsf{gather}}(\mathbf{EF}_i, \mathbf{B}_i),\, \mathbf{EF}_i,\, \mathbf{E}_i \right), 
\end{equation}

The Transformer encoder~\cite{vaswani2017attention} processes edge representations augmented with two special tokens, $\langle \mathrm{L} \rangle$ and $\langle \mathrm{E} \rangle$, where each edge embedding comprises context-aware features $\mathbf{h}_i^{EV}$ and endpoint descriptors. The context-aware features are computed via a Graph Convolutional Network (GCN)~\cite{kipf2016semi} operating on previously generated topology, effectively propagating contextual signals across the evolving structure. Following established practice~\cite{li2025dtgbrepgen}, we decompose each edge into two directional node embeddings to better capture cyclic or multi-way connectivity patterns. These refined embeddings are then fed into the encoder to produce contextualized edge representations.

The decoder autoregressively generates the edge connectivity sequence by attending to the encoded contextual embeddings. At each step, the index in $\mathbf{EV}_i^{\mathrm{seq}}$ is mapped to its corresponding contextual embedding, forming an ordered input sequence that preserves structural progression. Positional encodings are added to maintain sequential fidelity. Finally, a pointer network mechanism~\cite{vinyals2015pointer} is applied at each decoding step to produce a probability distribution over candidate edges $\mathbf{h}_i^{EV}$ and special tokens, allowing the model to predict the most likely next connection conditioned on the current state.

\subsubsection*{Level-2 Geometric Generation}
\label{sec: geometry_representation}

At this stage, we aim to model the conditional Level-2 geometry distribution $ p(\hat{\mathcal{G}}_2 \mid \hat{\mathcal{T}}_1, \hat{\mathcal{G}}_1, \hat{\mathcal{T}}_2) $. The face geometry $\mathcal{G}_{2,i}$ encompasses both face-level and vertex-level geometric structures, represented as $\mathcal{G}_{2,i} = \{\mathbf{F}_i, \mathbf{V}_i\}$.

For face geometry generation, we condition on multiple sources of topological and geometric context. 
Geometric cues include face bounding boxes $\mathbf{B}_i$ and edge-aware face features, obtained by aggregating edge features via $\operatorname{\mathsf{gather}}(\mathbf{E}_i, \mathbf{EF}_i)$.
Topological cues consist of the face-vertex adjacency $\mathbf{FV}_i$, which can be derived from the face-edge adjacency $\mathbf{FE}_i$, as well as $\mathbf{FE}_i$ itself, both of which are explicitly incorporated into the modeling process. 
In a manner analogous to low-level geometry generation, we employ a learnable encoding function $\Phi_F$ to fuse all geometric and topological information into a unified initial embedding:
\begin{equation}
    \mathbf{h}_i^{F} = \Phi_F\left( \operatorname{\mathsf{gather}}(\mathbf{FE}_i, \mathbf{E}_i), \mathbf{B}_i, \mathbf{FV}_i, \mathbf{FE}_i \right).
\end{equation}
This embedding incorporates the spatial position of each face, encoded in $\mathbf{B}_i$, the geometric features of incident edges obtained via a gather operation, and explicit topological adjacencies provided by $\mathbf{FV}_i$ and $\mathbf{FE}_i$, as illustrated in Fig.~\ref{fig:story}. The resulting representation is then fed into a Transformer-based diffusion architecture—specifically, the face denoising network $\mathcal{P}_F$—which iteratively refines the face coordinates $\mathbf{F}_i$ through denoising:
\begin{equation}
    \mathbf{F}_i = \mathcal{P}_F\big(\boldsymbol{\epsilon}_T;\, \mathbf{h}_i^F\big), \quad \text{where } \boldsymbol{\epsilon}_T \sim \mathcal{N}(\mathbf{0}, \mathbf{I}).
\end{equation}

Similarly, for vertex geometry generation, we condition on general topological and geometric contexts. We leverage vertex-to-face $\mathbf{VF}_i$ and vertex-to-edge $\mathbf{VE}_i$ incidence maps, and extract corresponding geometric features using the gather operation. The vertex embedding is formulated as:
\begin{equation}
    \mathbf{h}_i^V = \Phi_V\left( \operatorname{\mathsf{gather}}(\mathbf{VF}_i, \mathbf{B}_i), \operatorname{\mathsf{gather}}(\mathbf{VF}_i, \mathbf{F}_i), \mathbf{VF}_i, \mathbf{VE}_i \right).
\end{equation}
Notably, we exclude edge geometry $\mathbf{E}_i$ from the input. Directly incorporating edge geometry into the vertex generation pipeline leads to performance degradation. We hypothesize that this stems from redundant geometric signals, which may introduce conflicting gradients during inference. Such conflicts can destabilize the optimization process and compromise structural coherence in the generated output. The vertex embedding $\mathbf{h}_i^V$ is subsequently processed by the dedicated vertex denoising network $\mathcal{P}_V$ within the same diffusion framework to recover the final vertex coordinates $\mathbf{V}_i$:
\begin{equation}
    \mathbf{V}_i = \mathcal{P}_V\big(\boldsymbol{\epsilon}_T;\, \mathbf{h}_i^V\big), \quad \text{where } \boldsymbol{\epsilon}_T \sim \mathcal{N}(\mathbf{0}, \mathbf{I}).
\end{equation}

By jointly modeling global topology and local geometry within a diffusion-based generative framework composed of $\mathcal{P}_F$ and $\mathcal{P}_V$, our approach enables high-fidelity reconstruction and synthesis of complex 3D shapes with strong structural integrity.

\section{Experiments}
\begin{table*}[htbp]
\centering

\begin{tabular}{llS[table-format=1.3]S[table-format=2.2]S[table-format=1.3]S[table-format=2.2]S[table-format=2.2]S[table-format=2.2]S[table-format=2.2]S[table-format=1.2]}
\toprule
\multicolumn{2}{l}{\textbf{Dataset}} 
& {MMD $\downarrow$} & {COV $\uparrow$} & {JSD $\downarrow$} & {Valid $\uparrow$} & {Unique $\uparrow$} & {Novel $\uparrow$} & {CC $\uparrow$} & {MC $\uparrow$} \\
\midrule

\multirow{4}{*}{DeepCAD} 
& DeepCAD          & 1.021 & 72.43 & 1.623 & 60.50 & 91.27 & 93.74 & 10.51 & 1.81 \\
& BrepGen           & 0.874 & 71.13 & 1.584 & 69.27 & 99.08 & 99.54 & 9.30  & 0.86 \\
& DGTBrepGen        & 0.867 & \best{76.83} & 1.504 & 73.27 & 99.13 & 99.21 & 9.78  & 0.73 \\
& \textbf{HiDiGen} & \textbf{0.856} & 71.37 & \textbf{1.424} & \textbf{75.77} & \textbf{99.52} & \textbf{99.71} & \textbf{11.03} & \textbf{1.99} \\
\midrule

\multirow{3}{*}{ABC} 
& BrepGen           & 1.543 & 66.37 & 2.377 & 53.47 & 97.64 & 98.57 & 10.76 & 1.37 \\
& DGTBrepGen        & 1.455 & \best{68.20} & 2.024 & 63.53 & 98.86 & 99.02 & 11.45 & 1.06 \\
& \textbf{HiDiGen} & \textbf{1.414} & 65.43 & \textbf{1.845} & \textbf{65.77} & \textbf{99.18} & \textbf{99.67} & \textbf{11.27} & \textbf{1.48} \\
\bottomrule
\end{tabular}
\caption{\textbf{Quantitative comparison of shape generation methods on the DeepCAD and ABC datasets.} 
Lower values are better for MMD-CD and JSD (both scaled by $10^2$), while higher values are better for COV, Valid, Unique, Novel, CC, and MC. 
All metrics except CC and MC are reported as percentages. 
Best results are \best{bolded}.}
\label{tab:quantitative_results}

\end{table*}

\begin{figure*}[htbp]
    \makebox[\linewidth]{
        \includegraphics[width=1.0\linewidth]{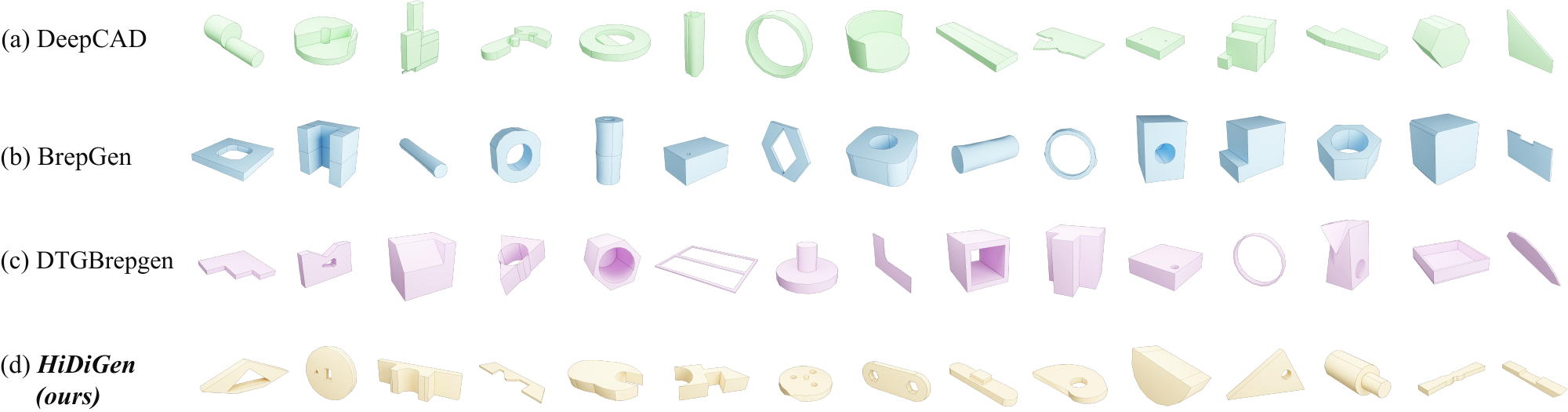}%
    }
\caption{Qualitative comparison of unconditioned B-rep models generated by our method and prior approaches, including DeepCAD ~\cite{wu2021deepcad}, BrepGen ~\cite{xu2024brepgen}, and DTGBrepGen ~\cite{li2025dtgbrepgen}, on the DeepCAD dataset. Our method generates more versatile and consistently watertight solid B-reps, demonstrating geometric fidelity and topological correctness.}
    \label{fig:uncond_compare}
\end{figure*}

In this section, we detail our experimental setup and present both qualitative and quantitative results on multiple datasets, covering both unconditional generation and point cloud generation. Additional results are provided in the Appendix.

\subsection{Experimental Setup}

\textbf{Datasets.} We evaluate generation performance on the DeepCAD~\cite{wu2021deepcad} and ABC~\cite{koch2019abc} datasets. To ensure consistency and complexity control~\cite{brepdiff,li2025dtgbrepgen}, we filter out models with more than 50 faces or more than 30 edges. After applying this preprocessing criterion, we obtain 60,478 valid samples from the DeepCAD dataset and 110,276 from the ABC dataset for our experiments.

\noindent\textbf{Implementation Details.} We employ a 4-layer Transformer architecture with 128-dimensional embeddings and 4 attention heads for both edge-face adjacency prediction and autoregressive edge-vertex generation. 
All models are trained with a learning rate of $5 \times 10^{-4}$ and a weight decay of $1 \times 10^{-6}$. 
For geometry generation, we adopt an 8-layer Transformer-based diffusion model with 512-dimensional embeddings and 8 attention heads, using the same optimization configuration. 
Training is conducted on a cluster equipped with 8 NVIDIA RTX 4090 GPUs.

\noindent\textbf{Evaluation Metrics.} Following the evaluation protocol of~\cite{xu2024brepgen}, we adopt a comprehensive set of metrics that includes both distribution-based and CAD-specific measures to thoroughly assess the fidelity, quality, and diversity of the generated 3D shapes.

Distribution Metrics include Coverage (COV), Minimum Matching Distance (MMD), and Jensen-Shannon Divergence (JSD), which measure the similarity between the generated and ground-truth distributions. 
In our implementation, COV and MMD are computed using Chamfer Distance (CD) over sampled surface points, following standard practice in point cloud comparison.
CAD metrics evaluate the geometric validity and uniqueness of the generated models, including Novel, Unique, and Valid ratios~\cite{jayaraman2022solidgen}. 
A model is considered \emph{Novel} if it does not appear in the training set, \emph{Unique} if it is generated only once across the evaluation set, and \emph{Valid} if it forms a watertight, manifold solid B-rep with correct topology and geometry. 

To further analyze the complexity of generated structures, we utilize two additional metrics  : Cyclomatic Complexity (CC)~\cite{contero2023quantitative} and Mean Curvature (MC). 
CC quantifies the topological complexity of a shape by counting the number of independent cycles in its wireframe graph representation, reflecting the richness of its connectivity. 
MC measures the average surface curvature over the shape's mesh, serving as an indicator of geometric detail and surface variation. 
Together, these metrics enable a holistic assessment of generation performance, encompassing distribution alignment, physical plausibility, diversity, and structural expressiveness.
\subsection{Unconditional Generation}

We compare our approach with three representative B-rep generation methods: DeepCAD~\cite{wu2021deepcad}, BrepGen~\cite{xu2024brepgen}, and DTGBrepGen~\cite{li2025dtgbrepgen}. For DeepCAD, we evaluate the reconstructed B-rep models from generated sketches and extrusion sequences. We randomly sample 3,000 generated models, and for each model, uniformly sample 2,000 points on its surface to compute distribution-based metrics. The CAD-specific metrics are computed directly on the 3,000 generated B-rep models. As shown in Table~\ref{tab:quantitative_results}, HiDiGen outperforms all baseline methods across nearly all evaluation metrics. Notably, HiDiGen achieves a significantly higher complexity score, indicating its superior ability to generate diverse and structurally rich designs. Qualitative results in Figure~\ref{fig:uncond_compare} further demonstrate that HiDiGen produces more realistic and geometrically accurate B-rep models compared to existing approaches, with better preservation of fine-grained topological and dimensional details.

\subsection{Conditional Generation}
\begin{figure}[htbp]
    \makebox[\linewidth]{
        \includegraphics[width=0.9\linewidth]{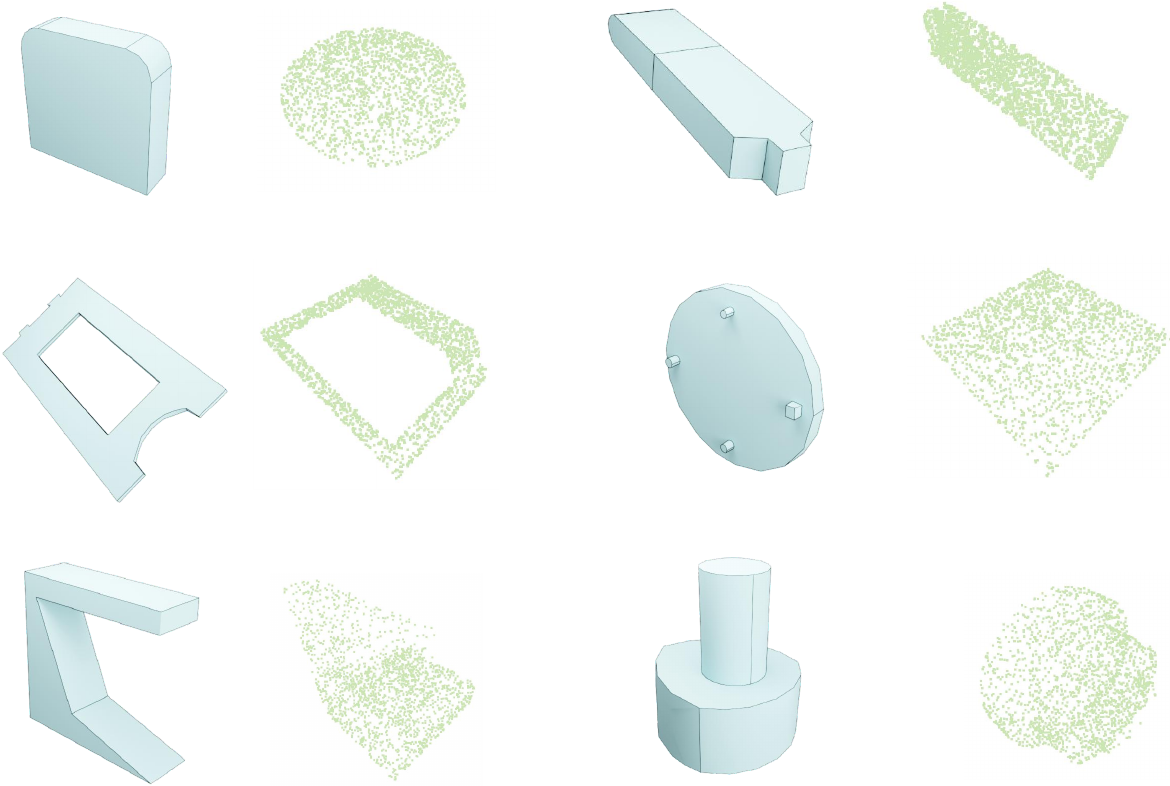}%
    }
    \caption{Point cloud conditioned B-rep generation.
 }
    \label{fig:pc}
\end{figure}

To enable point cloud conditioned generation, we adapt the Transformer architecture by first processing 2000 input points with PointNet++~\cite{qi2017pointnet++} to obtain a 512-dimensional global embedding that captures the geometric structure of the underlying shape. This geometry-aware contextual representation is then injected directly into the input token embeddings of the Transformer (e.g., $\mathbf{h}_i^{E}$, $\mathbf{h}_i^{EV}$), serving as a structural prior for the generative process, as illustrated in Fig.~\ref{fig:pipeline}. Qualitative results in Fig.~\ref{fig:pc} demonstrate that our model successfully preserves the coarse geometry and global layout of the input point cloud; however, the generated outputs occasionally lack fine grained details, indicating room for improvement in local texture and high frequency feature synthesis.

\subsection{Discussion}
\textbf{Ablation Study.} To demonstrate the effectiveness of leveraging prior geometric information in topology structure construction, we conduct an ablation study in which certain contextual cues are masked during the generation process. For Level-2 topology generation, the contextual information is given by $\mathbf{h}_i^{EV} = \Phi_{EV}\left( \operatorname{\mathsf{gather}}(\mathbf{EF}_i, \mathbf{B}_i),\, \mathbf{EF}_i,\, \mathbf{E}_i \right)$, which encodes the geometry of face bounding boxes adjacent to each edge as well as the edge’s own geometric features. Our ablation study considers two variants: (a) \textit{w/o}~$\mathbf{B}_i$, where the face bounding box information is masked, and (b) \textit{w/o}~$\mathbf{E}_i$, where the edge geometry is masked. Experimental results show that without geometric guidance, the validity rate of the generated B-rep models drops significantly. Quantitative results are reported in Table~\ref{tab:ablation}.

\begin{figure}[htbp]
    \makebox[\linewidth]{
        \includegraphics[width=1.0\linewidth]{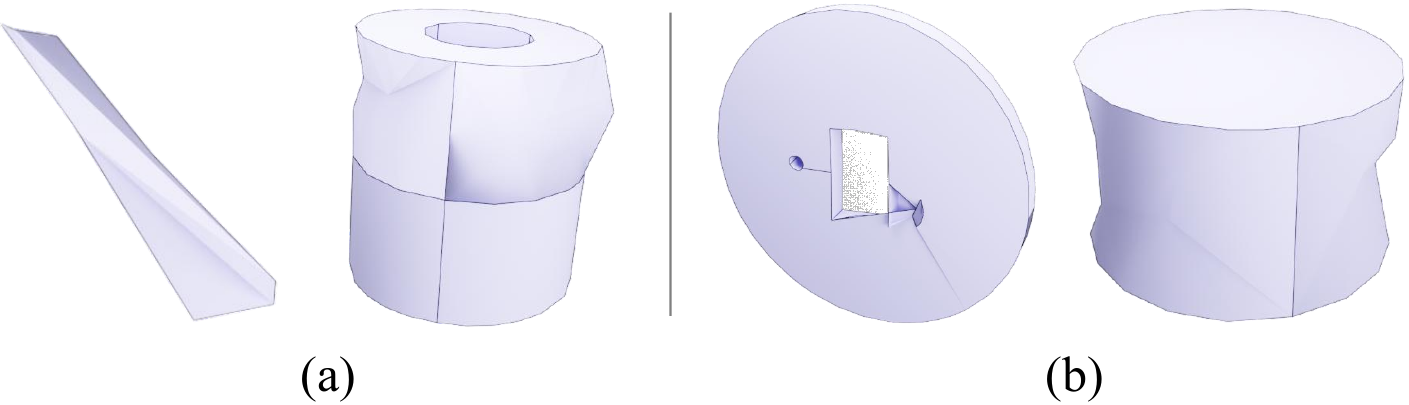}%
    }
    \caption{HiDiGen failure cases illustrating error propagation between topology and geometry: (a) erroneous topology yields incorrect geometry; (b) inaccurate geometry leads to spurious topological connections.
 }
    \label{fig:fail}
\end{figure}

\noindent\textbf{Failure Cases.} While the bidirectional coupling between topology and geometry in our pipeline improves overall output fidelity, the framework consists of two autoregressive stages followed by four diffusion stages, and errors introduced at any stage can propagate during inference. As shown in Fig.~\ref{fig:fail}, we observe two main failure modes: (a) an incorrect topological structure often leads to distorted geometry, as downstream diffusion stages struggle to recover plausible shapes from inconsistent structural priors; and (b) inaccurate geometric reconstruction can induce spurious topological connections, especially when surface and edge geometries are misaligned. These failures primarily stem from geometric inconsistencies between a surface and its incident edges or from erroneous edge connectivity in complex or ambiguous configurations.

\section{Conclusion}

\begin{table}[htbp]
\centering
\small 
\setlength{\tabcolsep}{5pt} 

\begin{tabular}{@{}llS[table-format=1.3]S[table-format=2.2]S[table-format=1.3]S[table-format=2.2]@{}}
\toprule
\multicolumn{2}{@{}l}{\textbf{Dataset}} 
& {MMD $\downarrow$} & {COV $\uparrow$} & {JSD $\downarrow$} & {Valid $\uparrow$} \\
\midrule

\multirow{3}{*}{DeepCAD} 
& \textit{w/o} $\mathbf{B}_i$           & 0.869 & 70.23 & 1.437 & 75.03 \\
& \textit{w/o} $\mathbf{E}_i$            & 0.871 & 70.87 & 1.441 & 73.27 \\
& \textbf{HiDiGen} & \textbf{0.856} & \textbf{71.37} & \textbf{1.424} & \textbf{75.77} \\
\midrule

\multirow{3}{*}{ABC} 
& \textit{w/o} $\mathbf{B}_i$            & 1.483 & 63.27 & 1.873 & 64.83 \\
& \textit{w/o} $\mathbf{E}_i$            & 1.579 & 61.40 & 1.948 & 62.80 \\
& \textbf{HiDiGen} & \textbf{1.414} & \textbf{65.43} & \textbf{1.845} & \textbf{65.77} \\
\bottomrule
\end{tabular}
\caption{Quantitative comparison of HiDiGen with ablation variants on DeepCAD and ABC datasets.}
\label{tab:ablation}

\end{table}

In this paper, we present HiDiGen, a hierarchical diffusion framework for B-rep generation that decouples topology and geometry in a structured, two-stage process. By first establishing a valid topological scaffold and then progressively refining geometric details under explicit connectivity constraints, HiDiGen generates novel, diverse, and topologically sound CAD models. Our method achieves state-of-the-art performance in B-rep generation, offering a promising step toward automatic and meaningful 3D design synthesis.

While HiDiGen demonstrates the necessity of hierarchical generation for B-rep modeling, where previously predicted structural and geometric cues are strategically leveraged to guide subsequent steps, the validity rate of the generated outputs remains suboptimal, indicating substantial room for improvement. 

Moreover, the current point cloud conditioned B-rep generation approach captures only the coarse shape of the input point cloud, leaving finer geometric details underexplored and highlighting clear opportunities for further development.

{
    \small
    \bibliographystyle{ieeenat_fullname}
    \bibliography{main}
}

\clearpage
\setcounter{page}{1}
\maketitlesupplementary
This supplementary material provides additional details and results to support the claims in the main paper. We first elaborate on the evaluation metrics used to assess geometric fidelity, topological validity, and generation diversity. We then present representative failure cases that highlight current limitations in jointly modeling topology and geometry , followed by extended qualitative results on the DeepCAD and ABC datasets.

\section{Details of Evaluation Metrics}

\label{sec:rationale}
In this section, we present more details about evaluation metric used in this paper.
\begin{itemize}
\item \textbf{Minimum Matching Distance (MMD-CD).} 
    This metric measures the average Chamfer Distance from each reference point cloud to its nearest neighbor in the generated set, reflecting the quality of the best-matching samples.

    \item \textbf{Coverage (COV-CD).} 
    COV-CD quantifies the diversity of the generated point clouds by computing the fraction of reference point clouds whose nearest neighbor in the generated set is unique. Higher coverage suggests that the model captures a broader range of the underlying data distribution.

    \item \textbf{Jensen–Shannon Divergence (JSD).}
    JSD evaluates the similarity between the occupancy distributions of real and generated point clouds on a discretized 3D grid. It provides a symmetric measure of distributional discrepancy, where a lower JSD implies better alignment between the synthetic and ground-truth data manifolds.

\item \textbf{Valid.} We report the percentage of generated B-reps that form watertight, non-manifold-free solids. A model is considered valid only if it satisfies fundamental topological constraints (e.g., each edge belongs to exactly two distinct faces and connects two distinct vertices) and passes geometric validation via a standard CAD kernel such as OpenCascade.

    \item \textbf{Novel.} Novelty measures the fraction of generated models that do not appear in the training set. We employ a hash-based identity check on the full B-rep structure. 

    \item \textbf{Unique.} Uniqueness quantifies intra-sample diversity by computing the proportion of generated models that occur only once in the output set. Using the B-rep hashing scheme, this metric reflects the model’s ability to avoid repetitive or mode-collapsed outputs.
\item \textbf{Average Number of Boundary Loops.}
    This metric counts the total number of topological boundary loops (i.e., \texttt{WIRE}s) across all faces of a CAD model and averages it over the dataset. Each loop corresponds to a connected boundary component of a face, including outer contours and inner holes. A higher value indicates more complex face topologies, often associated with intricate engineering features such as cutouts or fillets.

    \item \textbf{Average Mean Curvature.} 
    We compute the mean curvature $(k_1 + k_2)/2$ at uniformly sampled points on each face and average it across all faces and shapes. This scalar quantity captures the intrinsic smoothness and bending of surfaces: planar regions contribute zero curvature, while cylindrical or spherical regions yield non-zero values. The metric serves as a proxy for geometric complexity, with higher average curvature reflecting richer shape variation.
\end{itemize}

\section{Failure Analysis}

\begin{figure}[htbp]
    \centering
    \resizebox{\linewidth}{0.32\textheight}{\includegraphics{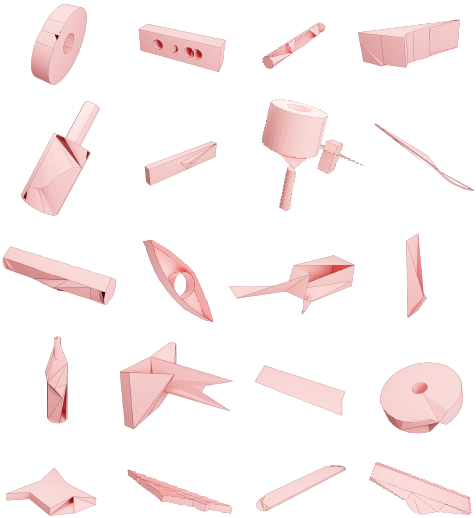}}
    \caption{Failure cases illustrating topological and geometric inconsistencies in generated B-reps.}
    \label{fig:morefail}
\end{figure}

Due to the intricate interplay between topological structure and geometric attributes in B-rep CAD generation, our approach employs four independent diffusion models and two autoregressive models. Although these models can leverage prior information during training, they may generate out-of-distribution predictions at inference time. This leads to two primary failure modes: (1) topological errors that propagate to produce geometrically invalid faces, and (2) inaccurate geometry that induces spurious topological connections. We illustrate representative failure cases in Fig.~\ref{fig:morefail}.

\section{Loss Functions}

Our framework is trained with three task-specific objectives corresponding to edge-face adjacency prediction, autoregressive edge-vertex sequence generation, and face geometry synthesis.

The edge-face module is trained as a variational autoencoder. Its loss combines a reconstruction term and a KL regularization term:
\begin{equation}
    \mathcal{L}_{\text{ef}} = \underbrace{\mathbb{E}\big[ -\log p(\mathbf{EF}_i^{seq} \mid \mathbf{z}) \big]}_{\text{reconstruction}} + \underbrace{D_{\mathrm{KL}}\big(q(\mathbf{z} \mid \mathbf{EF}_i^{seq}) \,\|\, \mathcal{N}(0, \mathbf{I}))\big)}_{\text{regularization}},
\end{equation}

where$\mathbf{EF}_i^{seq}$ denotes the ground-truth upper-triangular edge--face adjacency matrix, and $\mathbf{z}$ is the stochastic latent code. The reconstruction loss is implemented as cross-entropy over discrete adjacency entries, and the KL term admits the closed-form expression $-\frac{1}{2}\mathbb{E}[1 + \log\sigma^2 - \mu^2 - \sigma^2]$.

The edge-vertex generator is trained autoregressively via maximum likelihood. Given the ground-truth vertex sequence $\mathbf{v}_{1:T}$, the model predicts a categorical distribution over the next token at each step. The loss is the negative log-likelihood of the sequence, masked to exclude padding and special tokens, and normalized by the total number of valid prediction positions:

\begin{equation}
   \mathcal{L}_{\text{ev}} = -\frac{1}{\sum_{i,t} m_{i,t}} \sum_{i=1}^B \sum_{t=1}^{T-1} m_{i,t} \log p(v_{i,t+1} \mid v_{i,1:t}), 
\end{equation}

where $m_{i,t}$ indicates whether position $t$ in batch $i$ is a valid prediction target.

For geometry synthesis, we adopt a diffusion-based approach trained to predict noise added to face coordinates. The loss is the mean squared error between predicted and ground-truth noise, restricted to valid (non-padded) attributes:
\begin{equation}
    \mathcal{L}_{\text{geo}} = \frac{1}{|\mathcal{F}_{\text{valid}}|} \sum_{f \in \mathcal{F}_{\text{valid}}} \big\| \epsilon_{\theta}(f) - \epsilon \big\|_2^2,
\end{equation}

where $\epsilon$ is the sampled noise, $\epsilon_{\theta}(f)$ is the model’s prediction, and $\mathcal{F}_{\text{valid}}$ denotes the set of faces present in the input attributes.

All three losses are optimized jointly using AdamW with a learning rate of $5 \times 10^{-4}$ and weight decay of $1 \times 10^{-6}$.
\section{Additional Qualitative Results}
\begin{figure*}[htbp]
    \makebox[\linewidth]{
        \includegraphics[width=0.95\linewidth]{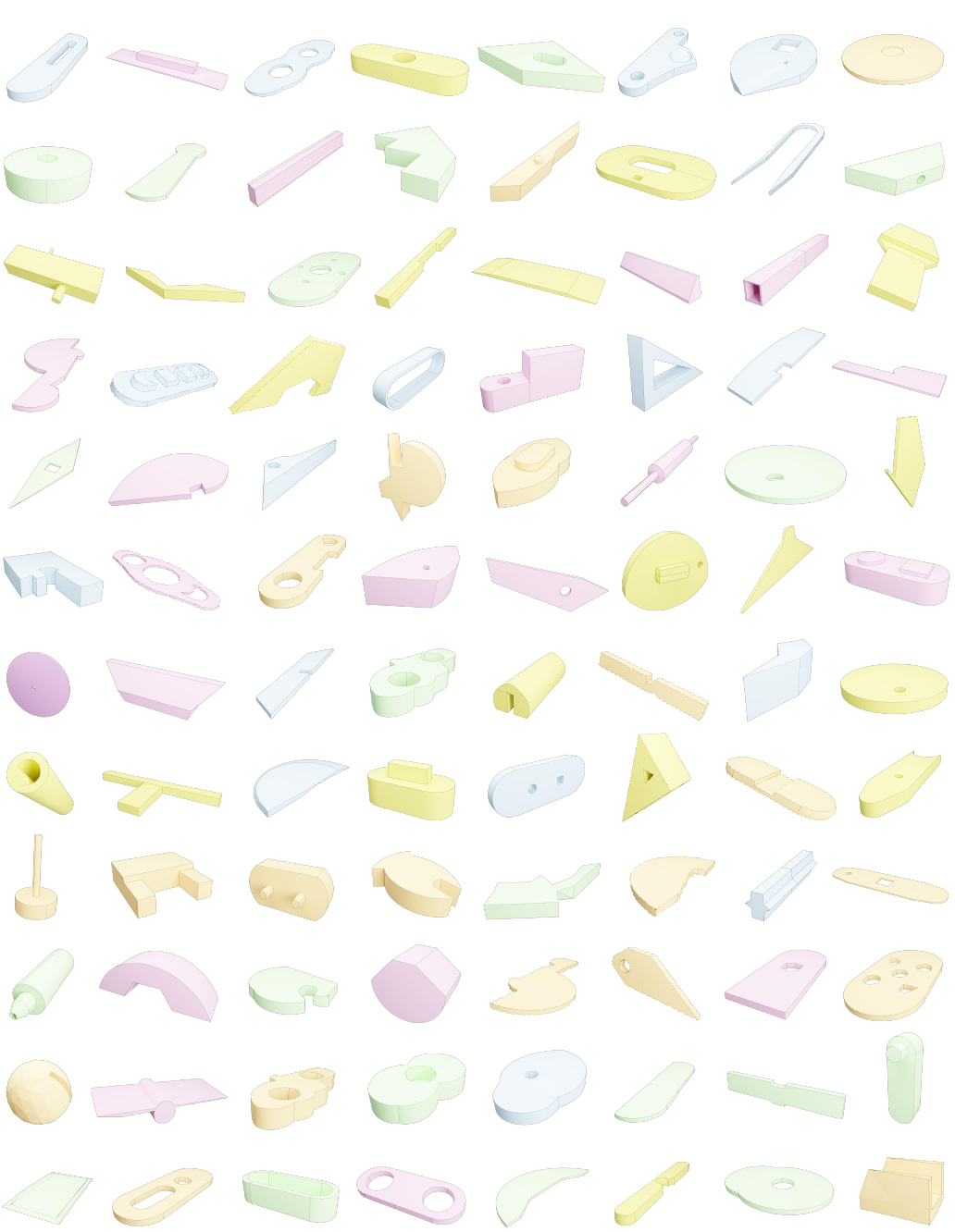}%
    }
\caption{Additional qualitative results of HiDiGen on the DeepCAD datasets.}
    \label{fig:moredeepcad}
\end{figure*}

Fig.~\ref{fig:moredeepcad} presents additional qualitative results of our method on the DeepCAD and ABC datasets.

\begin{figure*}[htbp]
    \makebox[\linewidth]{
        \includegraphics[width=0.95\linewidth]{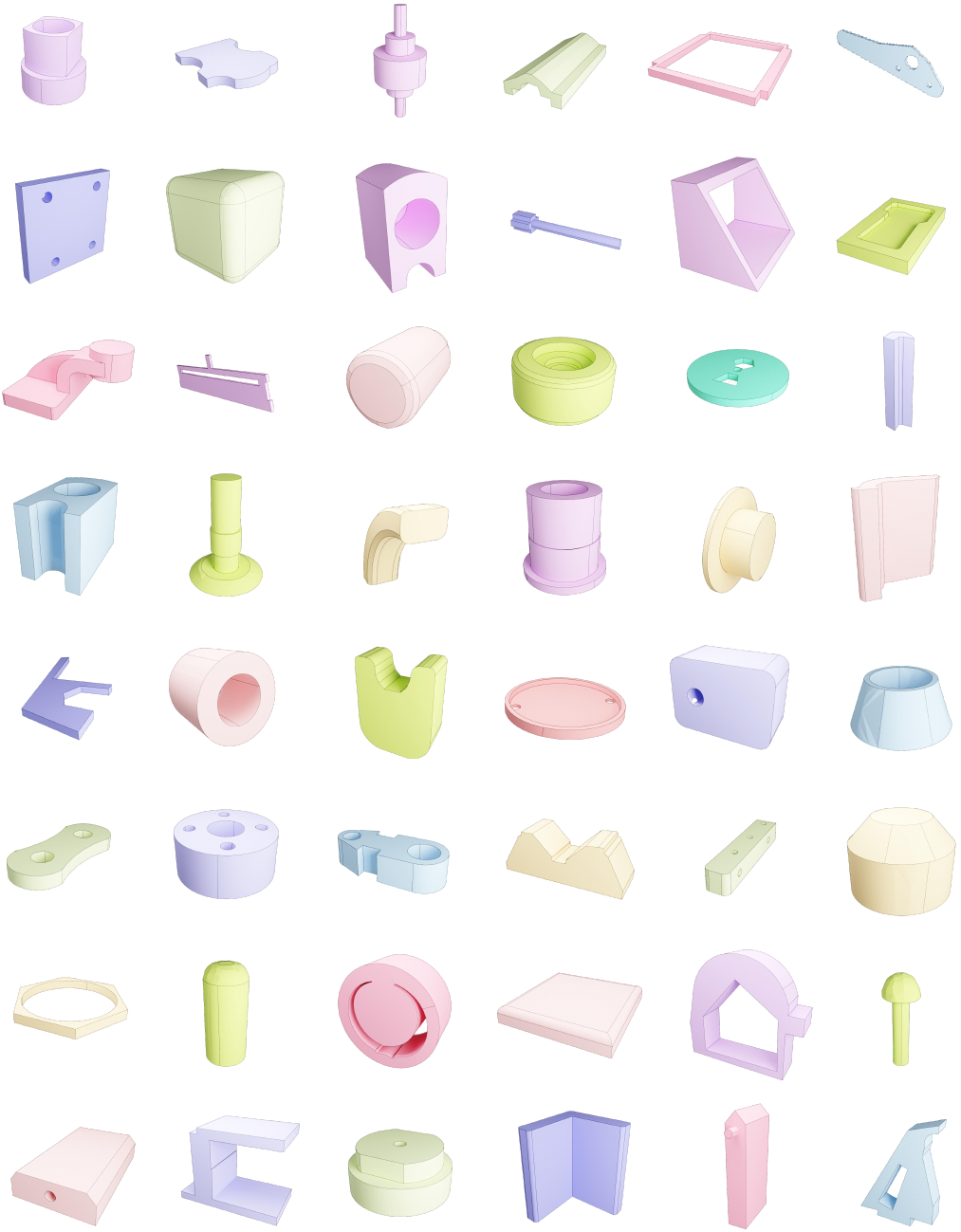}%
    }
\caption{Additional qualitative results of HiDiGen on the ABC datasets.}
    \label{fig:moreabc}
\end{figure*}

\end{document}